\documentclass{article}

\usepackage[sglblindworkshop,final]{neurips_2026}
\workshoptitle{2nd Workshop on Advances in Representation Learning for Earth Observation}

\makeatletter
\if@anonymous
  \newcommand{\acknowledgements}[1]{}
\else
  \newcommand{\acknowledgements}[1]{\section*{Acknowledgements}#1}
\fi
\makeatother

\usepackage[utf8]{inputenc}
\usepackage[T1]{fontenc}
\usepackage[hidelinks]{hyperref}
\usepackage{url}
\usepackage{booktabs}
\usepackage{amsmath}
\usepackage{amsfonts}
\usepackage{amssymb}
\usepackage{nicefrac}
\usepackage{microtype}
\usepackage{graphicx}
\usepackage{multirow}
\usepackage{float}
\usepackage{siunitx}
\usepackage{xcolor}
\usepackage{subcaption}

\usepackage{fancyhdr}

\AtBeginDocument{\setlength{\headheight}{15pt}}

\title{Ground-to-Satellite Knowledge Distillation for Crop Classification under Temporal and Geographic Shift}

\author{
  \textbf{Miro Miranda}\textsuperscript{1,*},
  \textbf{Rocco Sedona}\textsuperscript{1,*},
  \textbf{Valerio Marsocci}\textsuperscript{2},\\
  \textbf{Gabriele Cavallaro}\textsuperscript{1,3},
  \textbf{Claudia Paris}\textsuperscript{4}
  \\[0.5em]
  \small\normalfont
  \textsuperscript{1}J\"ulich Supercomputing Centre,
  Forschungszentrum J\"ulich, Germany\\
  \small\normalfont\textsuperscript{2}European Space Agency (ESA/ESRIN), Italy\\
  \small\normalfont\textsuperscript{3}University of Iceland, Iceland\\
  \small\normalfont\textsuperscript{4}University of Twente (ITC), the Netherlands
}

\begin{document}
\maketitle
\thispagestyle{plain}
\begingroup
\renewcommand{\thefootnote}{}
\footnotetext{*Equal contribution. }
\addtocounter{footnote}{-1}
\endgroup
\begin{abstract}
Continuous and up-to-date crop type mapping is essential for agricultural monitoring but is often hampered by limited reference data.
The growing availability of ground-level field photographs offers a promising way to address this challenge. When combined with satellite time series, these images provide complementary perspectives on the same parcel, capturing crop characteristics from distinctly different perspectives. However, field photographs are not always available for each satellite pixel at inference time, limiting their potential in scalable crop type mapping. We investigate whether ground photographs can provide privileged supervision for crop classification built on pretrained satellite embeddings. A photo teacher based on frozen CLIP features supervises classifiers trained on AlphaEarth and TESSERA representations; photographs are used only during training.
On a paired field photo and satellite imagery dataset with 17 crop classes, knowledge distillation consistently improves over satellite-only classification; for frozen embeddings the margin grows under temporal shift, reaching up to $+5.5$ macro-F1 points, and widens in geographic holdouts ($+3.7$ vs $+3.0$ in-distribution). These results show that ground-view supervision can complement satellite embeddings and improve their robustness without changing inference requirements.
\end{abstract}

\section{Introduction}
\label{sec:introduction}

Ground photographs offer a complementary perspective to satellite data and are becoming increasingly available for crop monitoring through both crowdsourcing platforms  \citep{garcin2021plntnetk} and field data collection protocols \citep{nakalembe2025helmets, dandrimont2022lucas}. For instance, the Land Use/Cover Area Frame survey (LUCAS) pairs georeferenced survey points across Europe with ground photographs, providing fine-grained crop characteristics such as leaf shape, structure, and texture that are difficult to capture from satellite data alone \citep{dandrimont2022lucas}. Such information can remain informative when satellite signature change across space or time and thus contributes to robust crop type mapping \citep{abbas2025multimodal}. However, the drawback is obvious: ground photographs are limited to surveyed locations and are rarely available on demand, which makes them impractical when rapid, scalable inference is required. This is a learning-using-privileged-information setting \citep{vapnik2009lupi}, for which Knowledge Distillation (KD) is a practical implementation \citep{hinton2015kd,lopezpaz2016unifying}.
We hypothesize that paired ground photographs provide a complementary semantic view of crop type that can improve robustness when scaling across regions and years \citep{wang2025unsupervised}. From an operational perspective, crop type maps require frequent updating to account for the highly dynamic nature of agricultural systems. \\
Recent advances in geospatial foundation-models (GFMs) offer the potential to streamline data preparation and substantially reduce Earth Observation (EO) data requirements by leveraging frozen embeddings for continuous mapping. However, limited reference data remains a key constraint, as classifiers trained on labels from specific geographic and temporal domains can suffer from distribution shifts, leading to misclassification \citep{wang2025unsupervised}. Such shifts arise because satellite observations are influenced not only by crop characteristics but also by climate, phenology, land management, and acquisition conditions \citep{wang2025unsupervised}, causing the same crop to exhibit different satellite signatures across domains. Ground photographs provide a complementary view of the same parcel that is less affected by these factors. Distilling the ground view into the classifier may therefore anchor predictions to crop-relevant cues that are more consistent across geographic and temporal domains. This is particularly relevant when classifiers are trained on frozen GFM embeddings, where the downstream model has limited ability to compensate for region- or time-specific biases encoded in the representation. \\
Building on the hypothesis that ground photographs provide complementary semantic information about crops, we investigate whether this information can be transferred to satellite-based classifiers to improve their robustness across geographic and temporal domains. Because ground photographs are not consistently available across space and time, we use a ground photo teacher only during training. Ground photographs are encoded with CLIP ViT-B/32 \citep{radford2021clip}, pretrained on approximately 400 million web-collected image--text pairs; a small photo teacher classifier is trained per seed and then frozen. Satellite students learn from the labels together with the teacher's logits and embeddings. At inference, the student receives satellite data only. We keep the teacher and loss fixed while changing the satellite branch, considering: (i) frozen embeddings from AlphaEarth \citep{brown2025alphaearth} and TESSERA \citep{feng2025tessera}, (ii) sequence transformers trained from scratch on monthly satellite time series, and (iii) a partially fine-tuned TESSERA encoder.

To assess the effectiveness of the proposed multimodal framework, we make three controlled comparisons. First, under the same teacher and training budget, we compare the KD--Cross Entropy (CE) difference across satellite models based on frozen and from-scratch  representations. Second, we evaluate these models under temporal transfer (2018$\rightarrow$2022) and, for AlphaEarth, held-out ecoregions, i.e., geographically distinct ecological regions \citep{dinerstein2017ecoregion}. Third, we test partial  fine-tuning of TESSERA while explicitly quantifying the mismatch introduced by our monthly-composite processing of satellite data. Across the main frozen and from-scratch regimes, KD consistently improves in-domain macro-F1 across the satellite branches, by up to 5.1 points on frozen TESSERA, and the frozen-embedding margins grow further under temporal shift and in the Mediterranean geographic holdout.

\section{Method}
\label{sec:method}

\textbf{Photo Teacher.} A frozen CLIP ViT-B/32 encoder \citep{radford2021clip} maps ground photographs to a feature vector $v$. Each views is individually L2-normed and averaged to create a single 512-dimensional representation over the location.
For each seed, a small photo teacher head is trained on the training split and then frozen: a two-layer projection to the shared 128-dimensional embedding $e_t(v)$ and a one-hidden-layer classifier producing the crop logits $l_t(e_t(v))$. The CLIP encoder itself remains frozen; the head is trained because frozen CLIP provides neither the shared embedding nor crop-class logits, the two distillation targets. Teacher targets are cached only for the student-training points.\\

\textbf{Satellite Student.} Each satellite student maps the input $z$ to a shared dimensional embedding $e_s(z)$ and logits $l_s(e_s(z))$.

We use (a) a small MLP head on frozen GFM embeddings, mirroring the teacher: a two-layer projection (width 256, LayerNorm, GELU) to $e_s(z)$ and a one-hidden-layer classifier on concatenated $z$ and $e_s(z)$; (b) a sequence transformer trained from scratch on monthly composites; and (c) TESSERA v1.1 \citep{feng2025tessera}, unfreezing the last two transformer blocks per modality and the fusion reducer. For (c), the encoder and head learning rates are $10^-4$ and $10^-3$, respectively; earlier layers remain frozen. Following standard knowledge distillation \citep{hinton2015kd}, with class label $y$, temperature $T$, and cross-entropy CE, the loss is:

\begin{equation}
\mathcal{L}=\mathrm{CE}(l_s,y)
+\lambda_{\mathrm{KL}}T^2\,\mathrm{KL}\!\left(\sigma(l_t/T)\,\|\,\sigma(l_s/T)\right)
+\lambda_{\cos}\left(1-\cos(e_s,e_t)\right).
\label{eq:loss}
\end{equation}

The CE control uses an identical student, optimization, and class-weighted CE term, with both distillation weights set to zero. Thus each reported KD--CE contrast changes only the privileged supervision.
More implementation details are provided in Appendix \ref{app:details}.

\section{Experimental setup}
\label{sec:experiments}

\begin{figure}[!tb]
    \centering
    \includegraphics[width=.9\linewidth]{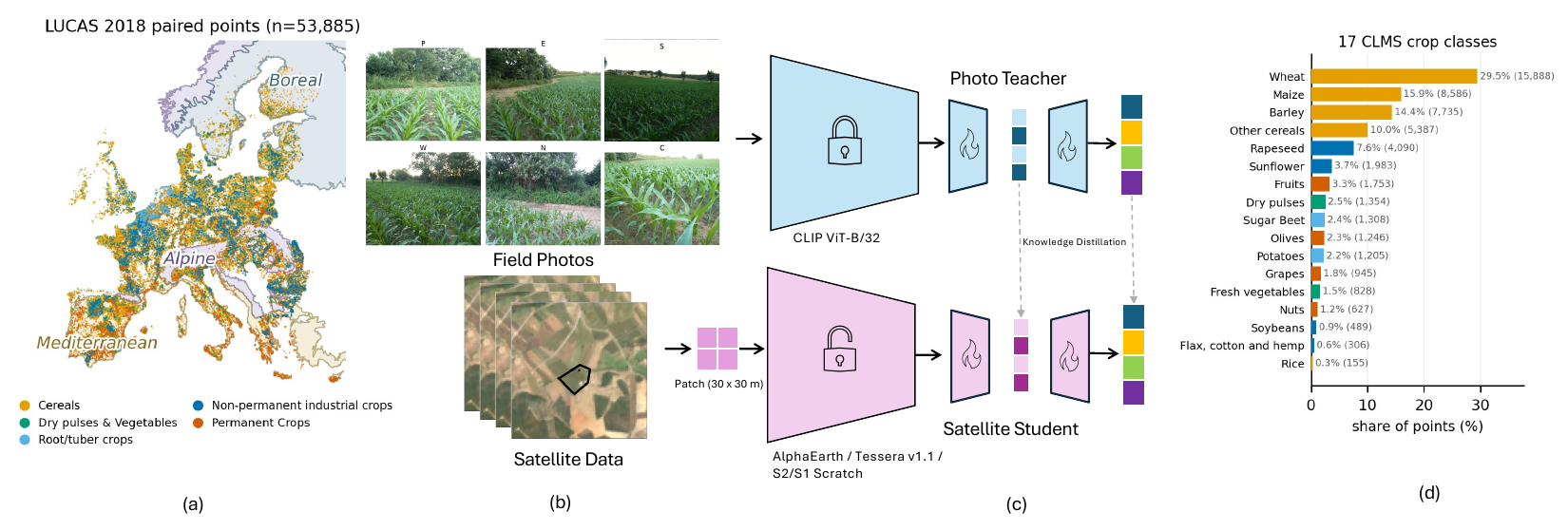}

    \caption{Dataset and model overview.
    (a) Spatial distribution of the selected agricultural LUCAS 2018 points,
    including the Alpine, Mediterranean, and Boreal held-out ecoregions.
    (b) Example of the Sentinel-2 annual composite (S2), the \textit{in-situ} photograph with the four cardinal views (E, W, N, S) and a close-up photograph (C). (c) KD workflow including photo teacher and satellite distilled student model.
    (d) Distribution of the 17 CLMS crop classes.}
    \label{fig:dataset_overview}
\end{figure}
\textbf{Data.}
To test the proposed approach, we use LUCAS 2018 survey points with ground photographs \citep{dandrimont2022lucas} for training, focusing on the 17 classes of the Copernicus High Resolution Layer Croplands (CLMS) products \citep{fahl2026europe}. The paired dataset contains 53{,}885 points with a label, photo, and time series of satellite data. The canonical split follows 70/15/15 for train/validation/test and is fixed across the main comparisons. The 2 \ensuremath{\times} 2 km LUCAS grid ensures spatially separated, non-overlapping observations, reducing potential spatial dependence between splits. Satellite inputs are either released 64-D AlphaEarth \citep{brown2025alphaearth} and TESSERA \citep{feng2025tessera} embeddings, or 12 monthly composites of Sentinel-2 (10 bands) and Sentinel-1 (VV/VH) over a $3\times3$ pixel patch ($30 $m$\times30$m) centered around the LUCAS point. Figure~\ref{fig:dataset_overview} provides an overview of the spatial distribution of crop types (a), Sentinel-2 composites, the \textit{in-situ} photograph, and the crop type distribution (d). \\
\textbf{Temporal transfer.}
Models trained on LUCAS 2018 are evaluated without adaptation on LUCAS 2022. The class mapping and normalization constants fitted on 2018 are reused, and geographical locations belonging to the 2018 train or validation sets are excluded from the 2022 transfer set. The resulting transfer set contains 36{,}428 points with a satellite feature and one of the 17 classes, at locations unseen during training. \\
\textbf{Geographic transfer.} Geographic transfer is evaluated entirely within 2018, isolating spatial from temporal domain shift. EEA ecoregions define fixed geographic test domains \citep{eea2016biogeo}: Mediterranean, Alpine, and Boreal are held out in turn, while the remaining 2018 points are split into training and validation sets. CE and KD are paired on the same split. Our input construction does not follow TESSERA's inference pattern: irregular acquisitions are reduced to 12 monthly composites, missing months become mean-valued observations with a valid mid-month day-of-year, and orbit-merged S1 uses averaged ascending/descending normalization statistics. A linear probe scores $36.2\pm1.6$ macro-F1 on embeddings from this frozen monthly-input path and $43.6\pm1.6$ on the released TESSERA embeddings, a $-7.4$ point difference. We therefore treat the released embedding and our monthly-input setting as separate regimes rather than using their difference in macro-F1 as an estimate of the effect of fine-tuning. The reported main experiments use five seeds. Macro-F1 is computed on the same evaluation set for each paired CE/KD run, using the fixed 17-label taxonomy.

\section{Results and Discussion}
\label{sec:results}
Table~\ref{tab:encoders} shows a consistent in-domain improvement of 2.6--5.1 macro-F1 points for the frozen and from-scratch students, largest on frozen TESSERA. Partial TESSERA fine-tuning (Sec.~\ref{sec:method}) gives the best absolute result, 72.1 macro-F1, with up to $+2.4$ macro-F1 over the teacher,  but the paired KD margin is smaller at $+0.8$ macro-F1. Because this branch uses the monthly input path above, this margin is not a controlled estimate of the effect of fine-tuning alone.

\begin{table}[t]
  \caption{Macro-F1 (\%), mean over five seeds, for matched CE and KD students. All students are trained on 2018 only; the transfer block evaluates them on the 2022 survey (unseen locations; Appendix~\ref{app:details}). $\Delta$ is the KD$-$CE mean of per-seed paired differences, computed before rounding. $\Delta$T is the difference to teacher performance (69.7 macro-F1).}
  \label{tab:encoders}
  \centering
  \small
\begin{tabular}{lccccccc}
    \toprule
    & \multicolumn{4}{c}{2018 test} & \multicolumn{3}{c}{2018$\rightarrow$2022 transfer} \\
    \cmidrule(lr){2-5}\cmidrule(lr){6-8}
    Satellite branch & CE & $+$KD & $\Delta$ & $\Delta$T & CE & $+$KD & $\Delta$ \\
    \midrule
    Frozen AlphaEarth        & 63.5 & 66.5 & $+2.9$ & $-3.2$ & 52.6 & 56.0 & $+3.4$ \\
    Frozen TESSERA           & 66.7 & 71.7 & $+5.1$ &  $+2.0$ & 57.6  & 63.1 & $+5.5$ \\
    S2 SITS (scratch)        & 62.0 & 65.2 & $+3.2$ & $-4.5$ & 54.0 & 56.4 & $+2.4$ \\
    S1$+$S2 SITS (scratch)   & 66.1 & 68.7 & $+2.6$ & $-1.0$ & 57.5 & 58.7 & $+1.2$ \\
    TESSERA fine-tuned       & 71.2 & \textbf{72.1} & $+0.8$ & $+2.4$ & 63.0 & \textbf{63.6} & $+0.7$\\
    \bottomrule
\end{tabular}
\end{table}

\begin{figure}[t]

    \centering
    \includegraphics[width=.76\linewidth]{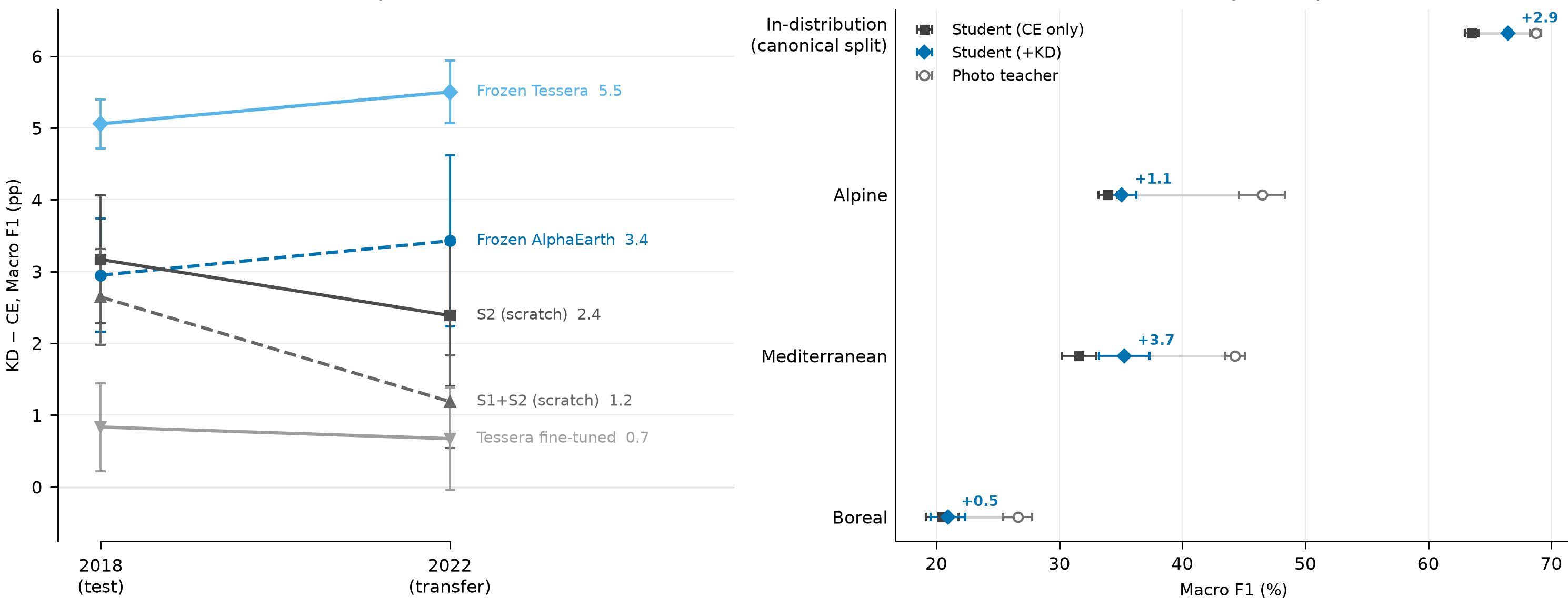}
    \hfill
    \caption{Left: Overall paired KD$-$CE macro-F1 margins per branch from 2018$\rightarrow$2022. Margins increase for frozen models, decrease for scratch models, and remain below one point for fine-tuned models. Right: AlphaEarth macro-F1 gains in held-out ecoregions. Only the Mediterranean exceeds the in-distribution gain ($+3.7$ vs. $+2.9$); Alpine gains $+1.1$, while Boreal shows unstable improvements ($+0.5$, SD $1.4$, mixed signs). Error bars indicate $\pm 1$ SD, paired for margins.}
    \label{fig:robustness}
    \vspace{-14pt}
\end{figure}

\par
\textbf{Transfer Experiments.}
Figure~\ref{fig:robustness} (left) shows the paired KD$-$CE margin for every branch under temporal transfer: the frozen margins grow ($+2.9$ to $+3.4$ macro-F1 points for AlphaEarth, $+5.1$ to $+5.5$ for TESSERA), the scratch margins shrink ($+3.2$ to $+2.4$ and $+2.6$ to $+1.2$), and the partially fine-tuned TESSERA branch stays below one point.
Geographic holdout (right) widens the margin in Mediterranean ($+3.7$ vs $+3.0$ in-distribution); Alpine is positive but smaller ($+1.1$), and Boreal is unstable ($+0.5$, sd 1.4, mixed signs); Mediterranean is thus the only clear geographic gain. \\
\textbf{Discussion.} Ground-view KD helps most clearly when the satellite representation is fixed, and its margin can grow under domain shift. Operationally, the photo teacher is needed only during training; deployment requires a satellite-only head on the same embedding product. KD also outperforms alternative cross-modal objectives such as contrastive learning (Appendix~\ref{app:CL}). The comparison has limits. First, TESSERA fine-tuning uses the dataset's monthly composites, for input consistency with the scratch transformers, whereas the frozen TESSERA model uses released embeddings produced through a different input path, a mismatch the $-7.4$ probe difference quantifies. Second, the geographic results are clear  in the Mediterranean holdout, Boreal shows how quickly a small holdout can become noisy. Lastly, the transfer performance depend on the class itself, especially for the frozen TESSERA (Appendix \ref{app:per_class}).

\section{Conclusion}
\label{sec:conclusion}
Ground photographs can provide privileged supervision for satellite crop classification without being required at inference. Their benefit is strongest for fixed pretrained satellite representations and persists, or increases, under temporal and geographic shift. Our ablations show that this gain is driven primarily by distillation of the teacher's predictive structure rather than by cross-modal feature alignment.
The gain is consistent across all satellite branches with frozen-embedding margins grow further under temporal shift and in the Mediterranean holdout. This knowledge-distillation framework is particularly powerful from an operational perspective: field observations can be used to improve satellite models without imposing field-data requirements on large-scale deployment. This is especially relevant for scaling agricultural mapping and monitoring, where satellite imagery enables broad and repeated coverage but ground observations are costly to collect.

\acknowledgements{

This work was supported by the European Space Agency (ESA) $\Phi$-Lab as part of the FAST-EO project, under Contract No. 4000143501/23/I-DT and the the AgroVision project, under contract No. 4000147606. The authors gratefully acknowledge the computing time granted on the supercomputer JUWELS Booster at J\"ulich Supercomputing Centre (JSC).}

\bibliographystyle{plainnat}
\bibliography{references}

\appendix

\section{Additional experimental details}
\label{app:details}
\paragraph{Sequence Transformer.} The time series of satellite images is encoded using a lightweight sequence transformer model. Monthly spectral features are linearly projected into a shared latent space. The sequence is processed by pre-normalized Transformer blocks with multi-head self-attention, feed-forward layers, residual connections, and dropout. Mean pooling across all months, followed by LayerNorm, produces a single sequence representation.

\paragraph{Photo Teacher} Precomputed photo features are mapped into a shared latent space using a two-layer projection head with LayerNorm and GELU activation. The resulting embedding is passed to an MLP classifier for crop type prediction. The trained photo teacher is frozen during student training and provides both embedding- and logit-level supervision.

\paragraph{Satellite student.} Precomputed satellite features are transformed into the teacher's latent space using an equivalent two-layer projection head. The projected embedding is concatenated with the original satellite feature vector and passed to an MLP classifier. Training combines supervised cross-entropy with logit- and embedding-level KD from the frozen photo teacher.

\paragraph{Main KD configuration.} The main KD runs use temperature $T=2$, logit-KD weight $1.0$, embedding-cosine weight $0.25$, and the same class-weighted CE term as the matched CE control. The student architecture, optimizer, split, and seed are paired within each KD--CE comparison.

\paragraph{Transfer controls.} The 2022 evaluation reuses the 2018 class map and scaler and excludes points used for 2018 training or validation. In the geographic experiments the held-out ecoregion is fixed as test data; only the train/validation partition of the remaining points changes with seed.

\paragraph{TESSERA comparison.} The frozen released-embedding experiment and the partially fine-tuned monthly-input experiment are deliberately reported as different regimes. A direct test of the effect of fine-tuning would require both CE and KD with the encoder frozen on the same monthly inputs.

\section{Temporal representation stability}
We further examined whether ground-view distillation changes the temporal stability of the learned satellite representation. This analysis was performed on the partially fine-tuned TESSERA student, using the encoder features before the projection and classification heads. We matched the 2018 and 2022 LUCAS surveys by point identifier and retained only locations observed in both years whose crop label was unchanged. This resulted in 1,046 repeated locations from 10 crop classes; locations whose crop label changed between surveys were excluded because a change in representation is expected in that case.

\begin{figure}[t]
    \centering
    \includegraphics[width=0.72\linewidth]{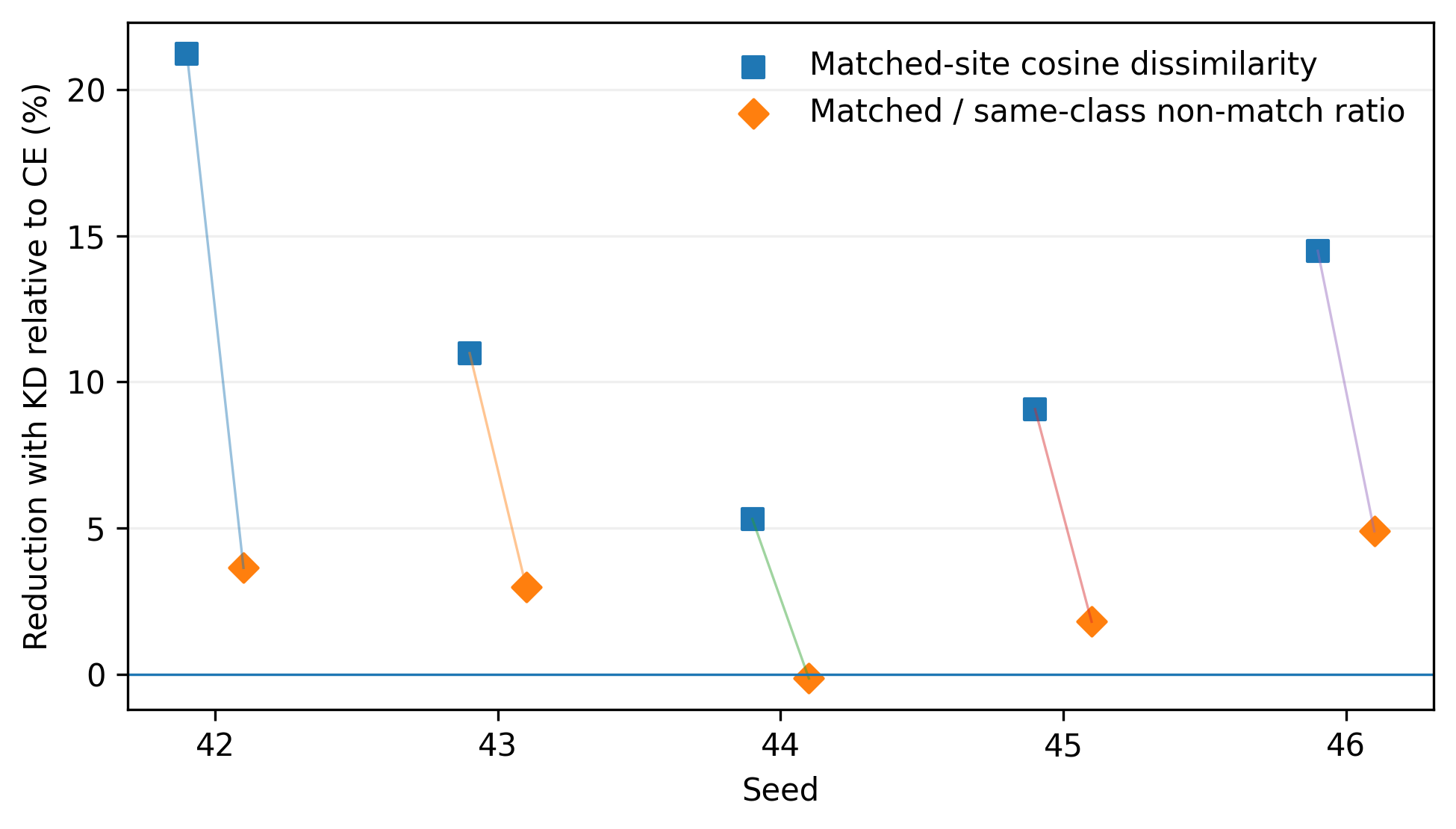}
    \caption{Temporal representation stability across five training seeds, computed on the 1{,}046 label-stable locations observed in both surveys. Squares: relative reduction from CE to KD, $\Delta_{\mathrm{site}}=100(D_{\mathrm{CE}}-D_{\mathrm{KD}})/D_{\mathrm{CE}}$, where $D$ is the matched-location dissimilarity of Eq.~\eqref{eq:temporal_cosine}, i.e.\ the cosine dissimilarity between the 2018 and 2022 representations of the same location, averaged within each crop class and then across classes. Diamonds: relative reduction $\Delta_{\mathrm{ratio}}=100(R_{\mathrm{CE}}-R_{\mathrm{KD}})/R_{\mathrm{CE}}$ of the matched-to-nonmatched ratio $R$ of Eq.~\eqref{eq:matched_nonmatched}, i.e.\ the same matched-location dissimilarity divided by the mean cross-year dissimilarity between a 2018 location and the other 2022 locations of its crop, so that a reduction in $R$ indicates location-specific stability beyond increased class compactness.}
    \label{fig:temporal_stability}
\end{figure}

For a retained location $i$, we quantify the change between years by the cosine dissimilarity
\begin{equation}
d_{\cos}\!\left(
\mathbf{h}^{2018}_{i},
\mathbf{h}^{2022}_{i}
\right)
=
1 -
\frac{
\mathbf{h}^{2018}_{i} \cdot \mathbf{h}^{2022}_{i}
}{
\left\|\mathbf{h}^{2018}_{i}\right\|_2
\left\|\mathbf{h}^{2022}_{i}\right\|_2
},
\label{eq:temporal_cosine}
\end{equation}
where $\mathbf{h}^{y}_{i}$ denotes the student representation of location $i$ in year $y$. We first average this quantity within each crop class and then macro-average over classes, such that frequent crops do not dominate the measurement. As shown in Fig.~\ref{fig:temporal_stability}, KD reduced the matched-location cross-year dissimilarity by 12.2\% on average across the five training seeds, with a reduction for every seed.

Because lower matched-location dissimilarity can also arise from increased within-class compactness, this quantity alone does not isolate a location-specific temporal effect.
We therefore used a same-class control. For each crop class $c$, the matched dissimilarity was compared with the mean cross-year dissimilarity between a 2018 location and all \emph{other} retained 2022
locations of the same crop:
\begin{equation}
R_c =
\frac{
\frac{1}{|\mathcal{I}_c|}
\sum_{i\in\mathcal{I}_c}
d_{\cos}\!\left(
\mathbf{h}^{2018}_{i},
\mathbf{h}^{2022}_{i}
\right)
}{
\frac{1}{|\mathcal{I}_c|}
\sum_{i\in\mathcal{I}_c}
\frac{1}{|\mathcal{I}_c|-1}
\sum_{\substack{j\in\mathcal{I}_c\\j\neq i}}
d_{\cos}\!\left(
\mathbf{h}^{2018}_{i},
\mathbf{h}^{2022}_{j}
\right)
},
\label{eq:matched_nonmatched}
\end{equation}
where $\mathcal{I}_c$ is the set of repeated, label-stable locations belonging to class $c$. The ratios $R_c$ are again macro-averaged across classes. KD reduced this matched-to-nonmatched ratio by 2.6\% on average, with lower values in four of five seeds. Thus, most of the reduction in cross-year dissimilarity co-occurs with increased intra-class compactness, while a smaller additional same-location stability effect remains. We therefore regard the result as consistent with the semantic-anchoring interpretation of ground-view distillation with strong regularization by the photo teacher's predictive structure.

\section{Contrastive Learning \& Modality Fusion}\label{app:CL}
We further tested alternative cross-modal objectives, such as contrastive learning (CL), including a supervised contrastive (SupCon) \citep{khosla2020supcon} and CLIP-style instance contrastive (InfoNCE) \citep{oord2018cpc,radford2021clip}. SupCon adds a supervised contrastive loss across modalities with positives defined by shared crop class. In contrast, InfoNCE adds an instance-contrastive loss with positives defined by the same site, aligning each satellite point with its own photo against the rest of the batch. We tested the CL approaches on the frozen AlphaEarth embeddings, coupled with the CLIP photo embeddings and compared against the AlphaEarth and photo probe. We further include the performance for combining both modalities. For the fusion model both modalities are concatenated at the input level and fed into the classifier.
Results for 2018 are shown in Table~\ref{tab:ablation_CL}. Importantly, the CL objectives do not improve over the matched CE control (per-seed paired differences are small and mixed in sign), in contrast to the KD objective repeated in the table ($+2.9$ macro-F1 paired over the CE student). The same dissociation holds within the distillation harness: adding only the logit-KL term to CE reproduces the full gain ($+2.9$ macro-F1, positive on every seed), whereas adding only the embedding-cosine term is null ($+0.2$ macro-F1, mixed signs). The CL objectives shape embeddings rather than predictive distributions, so their null result is consistent with the cosine-only control. In contrast, the macro-F1 of the photo probe is consistently about 6 pp higher than the CL objectives and both CE rows, which use the AlphaEarth embeddings only.
The combination of both modalities provides consistently the best results with approximately 8 pp higher macro-F1 than the photo probe model. This is expected, since this model receives complementary information that boosts crop classification performance but it cannot be used for inference in locations where no field photos are available.

\begin{table}[!tb]
\caption{Macro F1 and Weighted F1 for the frozen AlphaEarth CE baseline, contrastive objectives (InfoNCE, SupCon), photo probe, and fusion, mean $\pm$ std over the same five seeds, split, and evaluation protocol as Table~\ref{tab:encoders}. Photographs are used only during training for the satellite-only KD and contrastive models; the photo probe and fusion model require photographs at inference. The first two rows repeat the matched CE and KD students of Table~\ref{tab:encoders} as anchors. The CE control row is the in-harness baseline of the contrastive comparison (identical backbone, data, and optimizer as the CL rows); its head differs architecturally from the student, which accounts for the small offset between the two CE rows.}
\label{tab:ablation_CL}
\centering

\begin{tabular}{lrr}
\toprule
Model & MacroF1\% & WeightedF1\% \\
\midrule
Frozen AlphaEarth CE student [z] & 63.5 $\pm$ 0.6 & 67.0 $\pm$ 0.4 \\
+ KD [z] & 66.5 $\pm$ 0.3 & 68.6 $\pm$ 0.4 \\
CE control [z] & 63.8 $\pm$ 0.7 & 67.4 $\pm$ 0.5 \\
+ CLIP (InfoNCE) & 63.7 $\pm$ 0.7 & 67.2 $\pm$ 0.6 \\
+ SupCon & 63.4 $\pm$ 0.4 & 67.1 $\pm$ 0.4 \\
Photo probe [CLIP, v] & 69.7 $\pm$ 0.4 & 69.2 $\pm$ 0.3 \\
Fusion [z+v] & 77.7 $\pm$ 0.3 & 77.0 $\pm$ 0.3 \\
\bottomrule
\end{tabular}

\end{table}

\section{Per-class analysis}\label{app:per_class}

\begin{figure}[!tb]
    \centering
    \includegraphics[width=.65\linewidth]{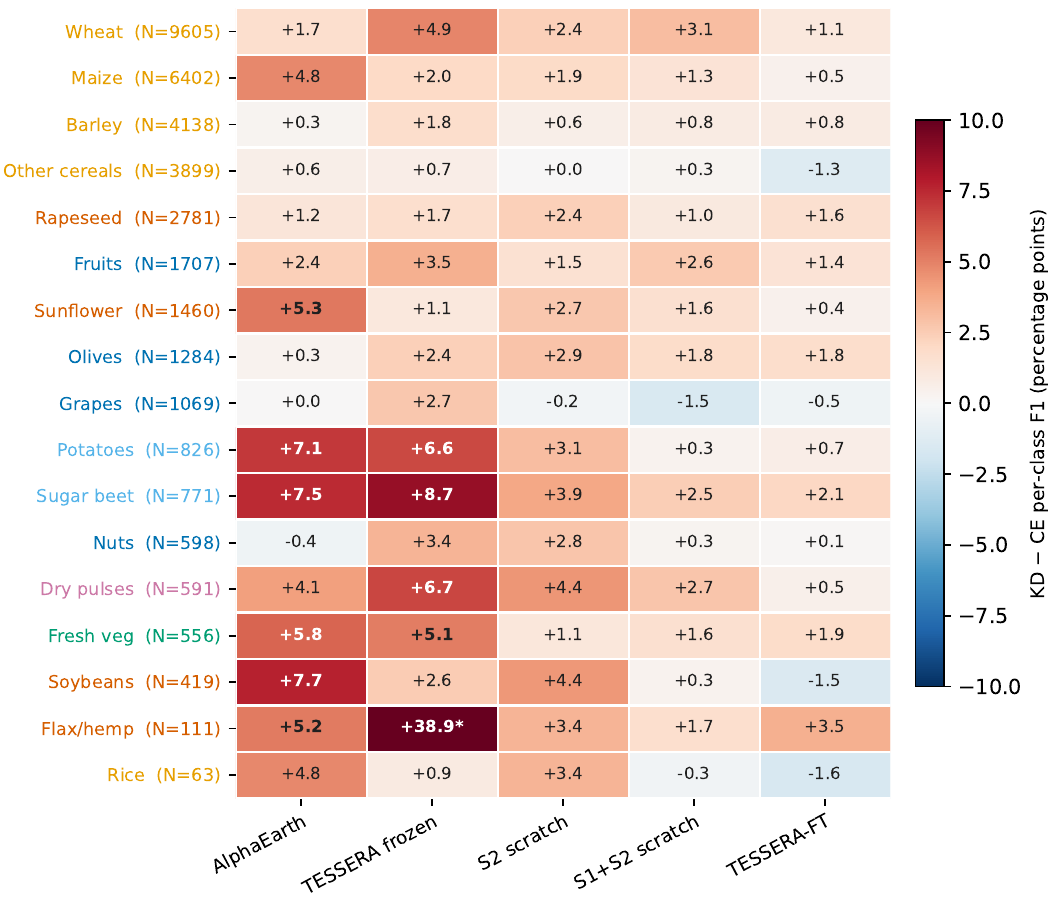}
    \caption{Heatmap for paired KD$-$CE macro-F1 margins per crop class for the temporal transfer (2022). Colorbar is cropped to 10.0 macro-F1 points.}
    \label{fig:per_class_tansfer}
\end{figure}

\begin{figure}[!tb]
    \centering
    \includegraphics[width=.7\linewidth]{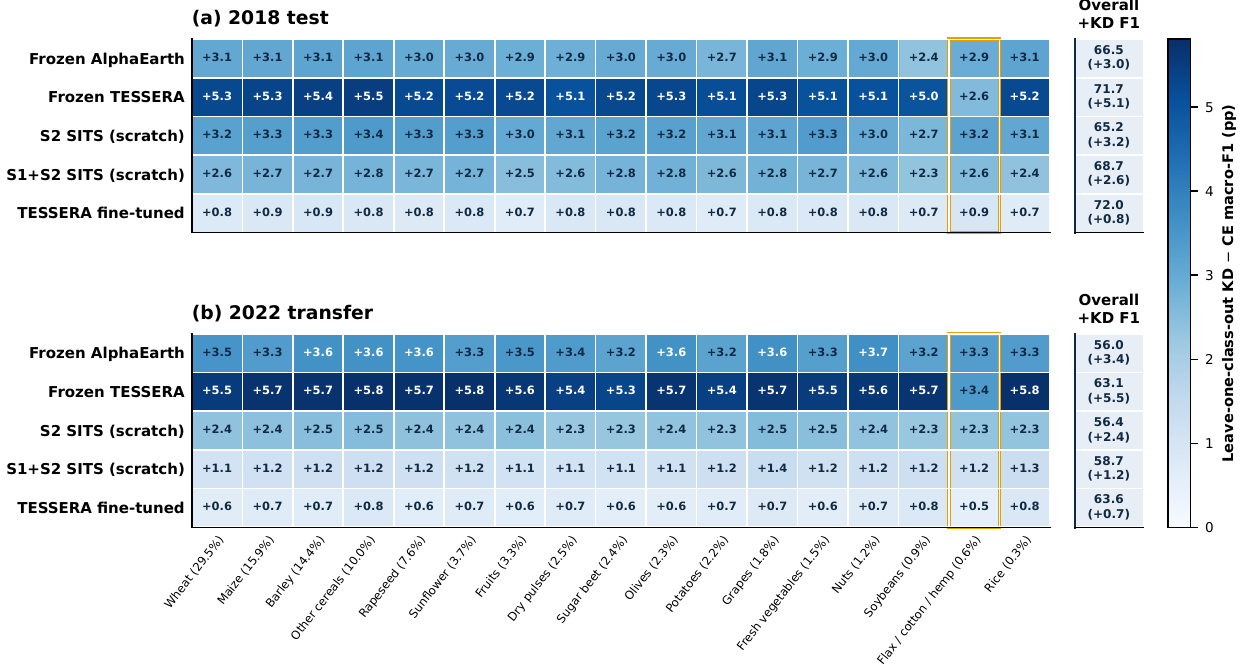}
    \caption{Heatmap for paired macro-F1 KD$-$CE margins fore a held-one-class-out validation over the model branches and yearsRre. Constantly stable performance gains, except for the frozen TESSERA.}
    \label{fig:retained_gain}
\end{figure}

We further show the per-class results of the marcro-F1 score on the 2018 test set in Table \ref{tab:per_class_f1_2018} and for the temporal transfer in Table \ref{tab:per_class_f1_2022}. Notably, KD improves crop classification for most classes, except individual cases such as rice (89.0 to 88.9) for the frozen AlphaEarth embeddings. Moreover, paired KD$-$CE macro-F1 margins per crop class for the temporal transfer is shown in Figure \ref{fig:per_class_tansfer}, evaluated at 2022. Notably, performance gains depends on crop type and model branch, with $+38.9$ macro-F1 points for Flax/Hemp on the frozen TESSERA model and $-1.6$ on Rice for the TESSERA fine-tuned branch.
To demonstrate that the reported KD gain is not only driven by single class results, we show the macro-F1 score for a held-one-class-out validation in Figure \ref{fig:retained_gain} with stable gains, except for frozen TESSERA. For frozen TESSERA under temporal transfer, excluding the combined flax, cotton, and hemp class reduces the gain from 5.5 to 3.4 percentage points.

\begin{table}[h!]
\caption{Per-class F1 (\%) on the 2018 test set, mean $\pm$ standard deviation over five seeds, for matched CE and KD students. Classes are sorted by 2018 test support.}
\label{tab:per_class_f1_2018}
\centering
\resizebox{\textwidth}{!}{
\begin{tabular}{lrrrrrrrrrr}
\toprule
 & \multicolumn{2}{c}{Frozen AlphaEarth} & \multicolumn{2}{c}{Frozen TESSERA}
 & \multicolumn{2}{c}{S2 SITS (scratch)}
 & \multicolumn{2}{c}{S1+S2 SITS (scratch)}
 & \multicolumn{2}{c}{TESSERA fine-tuned} \\
\cmidrule(lr){2-3} \cmidrule(lr){4-5} \cmidrule(lr){6-7}
\cmidrule(lr){8-9} \cmidrule(lr){10-11}
Class & CE & +KD & CE & +KD & CE & +KD & CE & +KD & CE & +KD \\
\midrule
    Wheat & 70.0 $\pm$ 0.9 & 71.3 $\pm$ 0.9 & 74.0 $\pm$ 0.2 & 76.0 $\pm$ 0.7 & 71.4 $\pm$ 1.3 & 73.7 $\pm$ 0.8 & 73.4 $\pm$ 2.2 & 76.3 $\pm$ 1.0 & 77.2 $\pm$ 1.1 & 78.3 $\pm$ 0.6 \\
    Maize & 87.8 $\pm$ 0.4 & 89.0 $\pm$ 0.2 & 89.0 $\pm$ 0.3 & 89.5 $\pm$ 0.2 & 88.6 $\pm$ 0.8 & 90.0 $\pm$ 0.3 & 89.9 $\pm$ 0.4 & 91.1 $\pm$ 0.3 & 91.4 $\pm$ 0.4 & 91.3 $\pm$ 0.4 \\
    Barley & 55.6 $\pm$ 1.1 & 55.6 $\pm$ 0.9 & 64.3 $\pm$ 1.0 & 63.6 $\pm$ 0.6 & 57.4 $\pm$ 1.3 & 57.9 $\pm$ 1.2 & 59.6 $\pm$ 0.6 & 60.8 $\pm$ 0.8 & 65.5 $\pm$ 0.8 & 64.8 $\pm$ 0.9 \\
    Other cereals & 46.6 $\pm$ 0.5 & 46.6 $\pm$ 0.4 & 51.4 $\pm$ 0.9 & 49.8 $\pm$ 1.2 & 44.1 $\pm$ 0.8 & 43.3 $\pm$ 1.1 & 48.6 $\pm$ 0.6 & 49.3 $\pm$ 0.9 & 53.5 $\pm$ 1.3 & 53.6 $\pm$ 1.0 \\
    Rapeseed & 77.4 $\pm$ 1.6 & 79.4 $\pm$ 0.6 & 80.6 $\pm$ 1.0 & 83.3 $\pm$ 0.3 & 79.0 $\pm$ 0.1 & 79.8 $\pm$ 0.5 & 82.1 $\pm$ 0.9 & 84.0 $\pm$ 0.5 & 82.1 $\pm$ 0.5 & 83.4 $\pm$ 0.4 \\
    Sunflower & 74.5 $\pm$ 1.3 & 77.4 $\pm$ 1.0 & 83.2 $\pm$ 1.5 & 86.2 $\pm$ 0.5 & 79.0 $\pm$ 1.3 & 80.0 $\pm$ 2.1 & 85.2 $\pm$ 1.0 & 86.1 $\pm$ 0.7 & 86.5 $\pm$ 1.1 & 87.2 $\pm$ 1.3 \\
    Fruits & 60.8 $\pm$ 1.0 & 64.9 $\pm$ 1.3 & 61.4 $\pm$ 1.2 & 64.0 $\pm$ 1.3 & 53.1 $\pm$ 1.7 & 58.4 $\pm$ 2.7 & 55.7 $\pm$ 3.7 & 60.7 $\pm$ 1.3 & 60.6 $\pm$ 0.9 & 63.6 $\pm$ 1.8 \\
    Dry pulses & 38.3 $\pm$ 1.5 & 42.6 $\pm$ 1.3 & 59.1 $\pm$ 3.3 & 64.0 $\pm$ 0.8 & 44.2 $\pm$ 1.0 & 48.7 $\pm$ 2.5 & 54.3 $\pm$ 2.6 & 58.3 $\pm$ 1.5 & 62.2 $\pm$ 1.0 & 62.6 $\pm$ 2.8 \\
    Sugar Beet & 81.8 $\pm$ 1.1 & 84.5 $\pm$ 1.6 & 84.2 $\pm$ 0.9 & 86.3 $\pm$ 0.7 & 83.6 $\pm$ 1.2 & 86.4 $\pm$ 1.6 & 86.9 $\pm$ 2.0 & 87.6 $\pm$ 1.4 & 87.7 $\pm$ 0.7 & 88.0 $\pm$ 0.8 \\
    Olives & 71.3 $\pm$ 1.9 & 74.3 $\pm$ 0.4 & 74.4 $\pm$ 1.7 & 76.0 $\pm$ 1.3 & 67.5 $\pm$ 1.4 & 69.7 $\pm$ 1.4 & 69.0 $\pm$ 1.2 & 69.1 $\pm$ 2.1 & 73.4 $\pm$ 1.9 & 74.0 $\pm$ 1.0 \\
    Potatoes & 60.3 $\pm$ 1.7 & 67.2 $\pm$ 1.6 & 68.0 $\pm$ 1.9 & 71.8 $\pm$ 1.3 & 62.2 $\pm$ 1.3 & 66.6 $\pm$ 1.8 & 69.8 $\pm$ 2.5 & 72.6 $\pm$ 1.1 & 75.6 $\pm$ 2.3 & 78.0 $\pm$ 1.3 \\
    Grapes & 60.3 $\pm$ 1.0 & 61.1 $\pm$ 1.1 & 65.7 $\pm$ 1.4 & 67.1 $\pm$ 2.0 & 52.0 $\pm$ 3.4 & 55.7 $\pm$ 0.9 & 56.5 $\pm$ 0.8 & 57.1 $\pm$ 0.8 & 62.0 $\pm$ 2.2 & 62.7 $\pm$ 2.6 \\
    Fresh vegetables & 30.6 $\pm$ 2.0 & 34.4 $\pm$ 1.6 & 39.3 $\pm$ 1.9 & 44.5 $\pm$ 0.5 & 30.6 $\pm$ 3.6 & 30.9 $\pm$ 3.1 & 33.7 $\pm$ 2.7 & 34.7 $\pm$ 2.9 & 39.1 $\pm$ 5.2 & 39.8 $\pm$ 4.0 \\
    Nuts & 41.3 $\pm$ 3.8 & 42.9 $\pm$ 2.1 & 39.3 $\pm$ 2.2 & 44.2 $\pm$ 4.0 & 33.9 $\pm$ 3.3 & 40.3 $\pm$ 3.0 & 38.3 $\pm$ 1.0 & 41.7 $\pm$ 1.5 & 41.5 $\pm$ 3.9 & 41.5 $\pm$ 2.7 \\
    Soybeans & 56.5 $\pm$ 3.7 & 69.1 $\pm$ 3.3 & 76.4 $\pm$ 1.7 & 82.5 $\pm$ 0.6 & 59.0 $\pm$ 3.8 & 70.2 $\pm$ 4.2 & 64.3 $\pm$ 2.7 & 71.9 $\pm$ 2.4 & 76.7 $\pm$ 1.2 & 78.4 $\pm$ 2.1 \\
    Flax, cotton and hemp & 77.8 $\pm$ 1.2 & 80.9 $\pm$ 2.0 & 35.8 $\pm$ 0.5 & 80.7 $\pm$ 2.6 & 70.4 $\pm$ 4.6 & 73.4 $\pm$ 1.2 & 73.2 $\pm$ 4.9 & 77.2 $\pm$ 2.6 & 85.8 $\pm$ 1.5 & 84.5 $\pm$ 2.4 \\
    Rice & 89.0 $\pm$ 3.8 & 88.9 $\pm$ 2.2 & 87.4 $\pm$ 2.9 & 90.1 $\pm$ 2.5 & 77.9 $\pm$ 5.6 & 82.7 $\pm$ 4.3 & 83.2 $\pm$ 4.6 & 90.1 $\pm$ 4.2 & 90.4 $\pm$ 1.8 & 93.1 $\pm$ 1.9 \\
\midrule
    Macro F1 & 63.5 $\pm$ 0.6 & 66.5 $\pm$ 0.3 & 66.7 $\pm$ 0.2 & 71.7 $\pm$ 0.2 & 62.0 $\pm$ 1.1 & 65.2 $\pm$ 1.0 & 66.1 $\pm$ 1.0 & 68.7 $\pm$ 0.6 & 71.2 $\pm$ 0.7 & 72.1 $\pm$ 0.5 \\
\bottomrule
\end{tabular}}
\end{table}

\begin{table}[h!]
\caption{Per-class F1 (\%) under 2018$\rightarrow$2022 temporal transfer, mean $\pm$ standard deviation over five seeds: the same trained students evaluated on the 2022 survey (unseen locations, train-year scaler and class map). Classes are sorted by 2018 test support.}
\label{tab:per_class_f1_2022}
\centering
\resizebox{\textwidth}{!}{
\begin{tabular}{lrrrrrrrrrr}
\toprule
 & \multicolumn{2}{c}{Frozen AlphaEarth} & \multicolumn{2}{c}{Frozen TESSERA}
 & \multicolumn{2}{c}{S2 SITS (scratch)}
 & \multicolumn{2}{c}{S1+S2 SITS (scratch)}
 & \multicolumn{2}{c}{TESSERA fine-tuned} \\
\cmidrule(lr){2-3} \cmidrule(lr){4-5} \cmidrule(lr){6-7}
\cmidrule(lr){8-9} \cmidrule(lr){10-11}
Class & CE & +KD & CE & +KD & CE & +KD & CE & +KD & CE & +KD \\
\midrule
    Wheat & 64.2 $\pm$ 0.4 & 65.8 $\pm$ 0.9 & 65.9 $\pm$ 1.2 & 70.8 $\pm$ 1.1 & 64.0 $\pm$ 1.4 & 66.4 $\pm$ 0.6 & 66.6 $\pm$ 2.5 & 69.7 $\pm$ 0.9 & 71.2 $\pm$ 1.5 & 72.3 $\pm$ 0.5 \\
    Maize & 78.5 $\pm$ 0.9 & 83.3 $\pm$ 0.2 & 86.3 $\pm$ 0.4 & 88.3 $\pm$ 0.2 & 83.8 $\pm$ 0.7 & 85.7 $\pm$ 0.9 & 85.8 $\pm$ 0.6 & 87.2 $\pm$ 0.2 & 88.1 $\pm$ 0.3 & 88.6 $\pm$ 0.3 \\
    Barley & 48.0 $\pm$ 1.9 & 48.3 $\pm$ 0.2 & 55.5 $\pm$ 0.5 & 57.3 $\pm$ 1.1 & 49.5 $\pm$ 1.0 & 50.1 $\pm$ 0.3 & 52.3 $\pm$ 0.8 & 53.1 $\pm$ 0.8 & 57.7 $\pm$ 1.4 & 58.5 $\pm$ 0.4 \\
    Other cereals & 42.2 $\pm$ 1.5 & 42.7 $\pm$ 1.9 & 47.6 $\pm$ 1.6 & 48.3 $\pm$ 1.5 & 41.9 $\pm$ 1.7 & 41.9 $\pm$ 1.8 & 46.0 $\pm$ 1.1 & 46.2 $\pm$ 0.5 & 50.3 $\pm$ 1.4 & 49.0 $\pm$ 1.9 \\
    Rapeseed & 68.4 $\pm$ 1.1 & 69.6 $\pm$ 0.4 & 71.8 $\pm$ 0.9 & 73.5 $\pm$ 0.5 & 67.8 $\pm$ 1.2 & 70.1 $\pm$ 0.8 & 73.3 $\pm$ 0.2 & 74.2 $\pm$ 0.1 & 72.7 $\pm$ 0.5 & 74.3 $\pm$ 0.5 \\
    Sunflower & 45.7 $\pm$ 2.2 & 51.0 $\pm$ 3.5 & 73.2 $\pm$ 1.4 & 74.2 $\pm$ 1.1 & 65.7 $\pm$ 1.3 & 68.4 $\pm$ 1.6 & 73.5 $\pm$ 0.9 & 75.1 $\pm$ 1.5 & 75.8 $\pm$ 1.2 & 76.2 $\pm$ 2.5 \\
    Fruits & 59.9 $\pm$ 1.1 & 62.4 $\pm$ 1.1 & 53.9 $\pm$ 1.6 & 57.4 $\pm$ 1.0 & 54.3 $\pm$ 0.6 & 55.8 $\pm$ 1.6 & 54.2 $\pm$ 1.9 & 56.8 $\pm$ 1.3 & 58.4 $\pm$ 0.8 & 59.8 $\pm$ 0.8 \\
    Dry pulses & 20.8 $\pm$ 2.3 & 24.9 $\pm$ 2.2 & 38.3 $\pm$ 3.3 & 45.0 $\pm$ 1.1 & 29.2 $\pm$ 1.5 & 33.7 $\pm$ 1.1 & 37.3 $\pm$ 2.7 & 40.1 $\pm$ 2.0 & 41.8 $\pm$ 0.8 & 42.3 $\pm$ 2.4 \\
    Sugar Beet & 64.6 $\pm$ 5.2 & 72.1 $\pm$ 1.5 & 64.9 $\pm$ 5.5 & 73.6 $\pm$ 4.5 & 76.2 $\pm$ 2.1 & 80.1 $\pm$ 2.3 & 79.1 $\pm$ 3.2 & 81.6 $\pm$ 1.4 & 78.7 $\pm$ 1.3 & 80.8 $\pm$ 2.2 \\
    Olives & 70.0 $\pm$ 1.7 & 70.3 $\pm$ 1.4 & 67.4 $\pm$ 1.5 & 69.8 $\pm$ 1.8 & 63.6 $\pm$ 1.8 & 66.4 $\pm$ 1.0 & 64.0 $\pm$ 1.6 & 65.8 $\pm$ 0.7 & 68.4 $\pm$ 0.7 & 70.2 $\pm$ 1.1 \\
    Potatoes & 41.0 $\pm$ 2.0 & 48.1 $\pm$ 2.5 & 56.9 $\pm$ 2.8 & 63.5 $\pm$ 2.1 & 51.9 $\pm$ 1.5 & 55.0 $\pm$ 1.9 & 57.5 $\pm$ 1.7 & 57.8 $\pm$ 1.5 & 63.7 $\pm$ 1.6 & 64.4 $\pm$ 1.1 \\
    Grapes & 54.7 $\pm$ 3.0 & 54.8 $\pm$ 5.0 & 49.6 $\pm$ 4.7 & 52.3 $\pm$ 3.0 & 46.8 $\pm$ 3.1 & 46.6 $\pm$ 3.9 & 53.3 $\pm$ 1.9 & 51.8 $\pm$ 2.4 & 55.7 $\pm$ 3.5 & 55.2 $\pm$ 2.6 \\
    Fresh vegetables & 22.7 $\pm$ 1.6 & 28.5 $\pm$ 1.5 & 30.2 $\pm$ 1.6 & 35.3 $\pm$ 1.7 & 28.3 $\pm$ 1.0 & 29.4 $\pm$ 3.2 & 29.1 $\pm$ 2.0 & 30.7 $\pm$ 1.1 & 34.3 $\pm$ 1.7 & 36.1 $\pm$ 1.2 \\
    Nuts & 32.1 $\pm$ 2.8 & 31.6 $\pm$ 1.6 & 29.5 $\pm$ 3.4 & 32.9 $\pm$ 1.7 & 25.2 $\pm$ 1.6 & 28.0 $\pm$ 0.9 & 28.9 $\pm$ 1.4 & 29.2 $\pm$ 1.1 & 30.1 $\pm$ 0.9 & 30.2 $\pm$ 0.8 \\
    Soybeans & 34.3 $\pm$ 2.6 & 42.0 $\pm$ 1.8 & 62.3 $\pm$ 3.4 & 64.9 $\pm$ 1.8 & 41.2 $\pm$ 1.5 & 45.6 $\pm$ 3.1 & 45.2 $\pm$ 1.7 & 45.5 $\pm$ 4.2 & 60.2 $\pm$ 1.9 & 58.8 $\pm$ 1.7 \\
    Flax, cotton and hemp & 65.6 $\pm$ 3.8 & 70.7 $\pm$ 3.1 & 39.8 $\pm$ 0.6 & 78.7 $\pm$ 2.3 & 58.6 $\pm$ 3.6 & 61.9 $\pm$ 2.9 & 56.7 $\pm$ 1.8 & 58.4 $\pm$ 2.8 & 74.9 $\pm$ 2.6 & 78.4 $\pm$ 1.3 \\
    Rice & 81.7 $\pm$ 3.1 & 86.5 $\pm$ 2.1 & 85.9 $\pm$ 1.2 & 86.8 $\pm$ 1.6 & 70.3 $\pm$ 4.9 & 73.7 $\pm$ 3.1 & 75.3 $\pm$ 3.9 & 75.0 $\pm$ 3.0 & 88.1 $\pm$ 1.5 & 86.5 $\pm$ 1.6 \\
\midrule
    Macro F1 & 52.6 $\pm$ 1.2 & 56.0 $\pm$ 0.6 & 57.6 $\pm$ 0.7 & 63.1 $\pm$ 0.6 & 54.0 $\pm$ 1.0 & 56.4 $\pm$ 0.7 & 57.5 $\pm$ 0.7 & 58.7 $\pm$ 0.1 & 63.0 $\pm$ 0.6 & 63.6 $\pm$ 0.4 \\
\bottomrule
\end{tabular}}
\end{table}

\end{document}